\documentclass[10pt,twocolumn,letterpaper]{article}

\usepackage{cvpr}
\usepackage{times}
\usepackage{epsfig}
\usepackage{amsmath}
\usepackage{amssymb}
\usepackage{xcolor}
\usepackage{microtype}
\usepackage{subfigure}
\usepackage{graphics}
\usepackage{graphicx}
\usepackage{booktabs} 
\usepackage{multirow}
\usepackage{amssymb}
\usepackage[rightcaption]{sidecap}
\usepackage{dsfont}
\usepackage{tabularx}
\usepackage{array}

\newcolumntype{?}[0]{!{\vrule width 0.05mm}}

\usepackage[pagebackref=true,breaklinks=true,letterpaper=true,colorlinks,bookmarks=false]{hyperref}

\cvprfinalcopy

\newcommand{\ptr}{\mathrm{P}_{\mathrm{tr}}}
\newcommand{\pte}{\mathrm{P}_{\mathrm{te}}}
\newcommand{\pr}{\mathrm{P}}
\newcommand{\score}{\mathrm{s}}
\newcommand{\knownd}{d^*}
\newcommand{\Lagr}{\mathcal{L}}

\newcommand{\smallsec}[1]{\vspace{0.01in} \noindent {\bf #1.}}

\DeclareMathOperator*{\argmax}{arg\,max}

\newcommand{\myequation}{\begin{equation}}
\newcommand{\myendequation}{\end{equation}}
\let\[\myequation
\let\]\myendequation

\ifcvprfinal\fi
\usepackage{bbm}
\begin{document}

\title{Towards Fairness in Visual Recognition: Effective Strategies for Bias Mitigation}
\author{Zeyu Wang, Klint Qinami, Ioannis Christos Karakozis, 
Kyle Genova, \\ Prem Nair, Kenji Hata, Olga Russakovsky \\
Princeton University\\
%{\footnotesize \texttt{\{zeyuwang,kqinami,kgenova,olgarus\}@cs.princeton.edu} \texttt{\{yannis.karakozis,prem.q.nair,a.kenjihata\}@gmail.com}}
\texttt{\{zeyuwang, olgarus\}@cs.princeton.edu}
}

\maketitle
\thispagestyle{empty}

%%%%%%%%% ABSTRACT
\begin{abstract}
Computer vision models learn to perform a task by capturing relevant statistics from training data. It has been shown that  models  learn spurious age, gender, and race correlations when trained for seemingly unrelated tasks like activity recognition or image captioning. Various mitigation techniques have been presented to prevent models from utilizing or learning such biases. However, there has been little systematic comparison between these techniques. We design a simple but surprisingly effective visual recognition benchmark for studying bias mitigation. Using this benchmark, we provide a thorough analysis of a wide range of techniques. We highlight the shortcomings of  popular adversarial training approaches for bias mitigation, propose a simple but similarly effective alternative to the inference-time Reducing Bias Amplification method of Zhao et al., and design a domain-independent training technique that outperforms all other methods. Finally, we validate our findings on the attribute classification task in the CelebA dataset, where attribute presence is known to be correlated with the gender of people in the image, and demonstrate that the proposed technique is effective at mitigating real-world gender bias.  
\end{abstract}
\vspace{-5mm}
\section{Introduction}

Computer vision models learn to perform a task by capturing relevant statistics from training data. These statistics range from low-level information about color or composition (zebras are black-and-white, chairs have legs) to contextual or societal cues (basketball players often wear jerseys, programmers are often male). Capturing these statistical correlations is  helpful for the task at hand: chairs without legs are rare and programmers who are not male are rare, so capturing these dominant features will yield high accuracy on the target task of recognizing chairs or programmers. However, as computer vision systems are deployed at scale and in a variety of settings, especially where the initial training data and the final end task may be mismatched, it becomes increasingly important to both \emph{identify} and develop strategies for \emph{manipulating} the information learned by the model. 

\smallsec{Societal Context}  To motivate the work of this paper, consider one such example of social bias propagation: AI models that have learned to correlate activities with gender~\cite{bolukbasi2016man,Caliskan2017science,zhao_men_2017,anne2018women}. Some real-world activities are more commonly performed by women  and others by men. This real-world gender distribution skew becomes part of the data that trains models to recognize or reason about these activities.\footnote{Buolamwini and Gebru~\cite{gender-shades-2018} note that collecting a more representative training dataset should be the first step of the solution. That is true in the cases they consider (where people with darker skin tones are dramatically and unreasonably undersampled in datasets) but may not be a viable approach to cases where the  datasets accurately reflect the real-world skew.} Naturally, these models then learn discriminative cues which include the gender of the actors. In fact, the gender correlation may even become \emph{amplified} in the model, as Zhao et al.~\cite{zhao_men_2017} demonstrates. We refer the reader to e.g., \cite{algorithms-of-oppression} for a deeper look at these issues and their impact. 

\smallsec{Study Objectives and Contributions} In this work, we set out to provide an in-depth look at this problem of training visual classifiers in the presence of spurious correlations. We are inspired by prior work on machine learning fairness~\cite{zhang2018mitigating,zhao_men_2017,ryu2017improving,alvi2018turning} and aim to build a unified understanding of the proposed techniques. Code is available at \url{https://github.com/princetonvisualai/DomainBiasMitigation}.

We begin by proposing a simple but surprisingly effective benchmark for studying the effect of data bias on visual recognition tasks. Classical literature on mitigating bias generally operates on simpler (often linear) models~\cite{dwork2012fairness,zemel2013learning, leino2018feature}, which are easier to understand and control; only recently have researchers begun looking at mitigating bias in end-to-end trained deep learning models~\cite{ganin2015unsupervised,anne2018women,ryu2018inclusivefacenet, grover2019bias, wang2019balanced, kim2019learning, li2019repair, quadrianto2019discovering, wang2019racial, grover2019fair}. Our work helps bridge the gap, proposing an avenue for exploring mitigating bias in Convolutional Neural Network (CNN) models within a simpler and easier-to-analyze setting than with a fully-fledged black-box system. By utilizing dataset augmentation to introduce controlled biases, we provide simple and precise targets for model evaluation (Sec.~\ref{sec:cifar}).

Using this benchmark, we demonstrate that the presence of spurious bias in the training data severely degrades the accuracy of current models, even when the biased dataset contains strictly more information than an unbiased dataset. We then provide a thorough comparison of existing methods for bias mitigation, including domain adversarial training~\cite{tzeng2015simultaneous,ryu2017improving,alvi2018turning}, Reducing Bias Amplification~\cite{zhao_men_2017}, and domain conditional training similar to~\cite{ryu2018inclusivefacenet}. To the best of our knowledge, no such comparison exists currently as these methods have been evaluated on different benchmarks under varying conditions and have not been compared directly. We conclude that a domain-independent approach inspired by~\cite{dwork2012fairness} outperforms more complex competitors (Sec.~\ref{sec:cifar-baselines}). 

Finally, we validate our findings in more realistic settings. We  evaluate on the CelebA~\cite{liu2015faceattributes} benchmark for attribute recognition in the presence of gender bias (Sec.~\ref{sec:real-world}). We demonstrate that our domain-independent training model successfully mitigates real-world gender bias.

\section{Related Work} 

\smallsec{Mitigating Spurious Correlation} Recent work on the effects of human bias on machine learning models investigates two challenging problems: identifying and quantifying bias in datasets, and mitigating its harmful effects. In relation to the former, \cite{buda_systematic_2017, liu2009exploratory} study the effect of class-imbalance on learning, while \cite{zhao_men_2017} reveal the surprising phenomenon of bias amplification. Additionally, recent works have shown that ML models possess bias towards legally protected classes \cite{beautycontest,gender-shades-2018,bolukbasi2016man,Caliskan2017science,madras2018learning,creager2019flexibly}. Our work complements these by presenting a dataset that allows us to isolate and control bias precisely, alleviating the usual difficulties of quantifying bias.

On the bias mitigation side, early works investigate techniques for simpler linear models \cite{khosla2012undoing, zemel2013learning}. Our constructed dataset allows us to isolate bias while not simplifying our architecture. More recently, works have begun looking at more sophisticated models. For example, \cite{zhao_men_2017} propose an inference update scheme to match a target distribution, which can remove bias. \cite{ryu2018inclusivefacenet} introduce InclusiveFaceNet for improved attribute detection across gender and race subgroups; our discriminative architecture is inspired by this work. Conversely, \cite{dwork2018decoupled} propose a scheme for decoupling classifiers, which we use to create our domain independent architecture. The last relevant approach to bias mitigation for us is adversarial mitigation \cite{alvi2018turning,zhang2018mitigating,edwards2016censoring,ganin2015unsupervised}. Our work uses our novel dataset to explicitly highlight the drawbacks, and offers a comparison between these mitigation strategies that would be impossible without access to a bias-controlled environment.

\smallsec{Fairness Criterion}
Pinning down an exact and generally applicable notion of fairness is an inherently difficult and important task. Various fairness criteria have been introduced and analyzed, including demographic parity \cite{kilbertus2017avoiding,zhang2018mitigating}, predictive parity \cite{gajane2017formalizing}, error-rate balance \cite{hardt2016equality}, equality-of-odds and equality-of-opportunity \cite{hardt2016equality}, and fairness-through-unawareness \cite{Pedreshi:2008:DDM:1401890.1401959} to try to quantify bias. Recent work has shown that such criteria must be selected carefully; \cite{hardt2016equality} prove minimizing error disparity across populations, even under relaxed assumptions, is equivalent to randomized predictions; \cite{hardt2016equality} introduce and explain the limitations of an `oblivious' discrimination criterion through a non-identifiability result;  \cite{Pedreshi:2008:DDM:1401890.1401959} demonstrate that ignoring protected attributes is ineffective due to redundant encoding; \cite{dwork2012fairness} show that demographic parity does not ensure fairness. We define our tasks such that test accuracy directly represents model bias. %adopt the criteria of model utility per domain, weighted mean average precision, and a simple notion of inference bias, defined in Equation~\ref{eq:bias}.

\smallsec{Surveying Evaluations}
We are inspired by previous work which aggregate ideas, methods and findings to provide a unify survey of a subfield of computer vision~\cite{imagenetTransferLearning,russakovskyiccv13,Sigurdsson_2017,hoiemODerrors}. For example, \cite{torralba_unbiased_2011} surveys relative dataset biases present in computer vision datasets, including selection bias (datasets favoring certain types of images), capture bias (photographers take similar photos), category bias (inconsistent or imprecise category definitions), and negative set bias (unrepresentative or unbalanced negative instances).  We continue this line of work for bias mitigation methods for modern visual recognition systems, introducing a benchmark for evaluation which isolates bias, and showing that our analysis generalizes to other, more complex, biased datasets.

\section{A Simple Setting for Studying Bias}
\label{sec:cifar}
We begin by constructing a novel benchmark for studying bias mitigation in visual recognition models. This setting makes it possible to demonstrate that the presence of spurious correlations in training data severely degrades the performance of current models, even if learning such spurious correlations is sub-optimal for the target task. 

\smallsec{CIFAR-10S Setup} To do so, we design a benchmark that erroneously correlates target classification decisions (what object category is depicted in the image) with an auxiliary attribute (whether the image is color or grayscale).

We introduce CIFAR-10 Skewed (CIFAR-10S), based on CIFAR-10 \cite{krizhevsky_learning_2009}, a dataset with 50,000 $32 \times 32$ images evenly distributed between 10 object classes. In CIFAR-10S, each of the 10 original classes is subdivided into two new domain subclasses, corresponding to color and grayscale domains within that class. Per class, the 5,000 training images are split 95\% to 5\% between the two domains; five classes are 95\% color and five classes are 95\% grayscale. The total number of images allocated to each domain is thus balanced. For testing, we create two copies of the standard CIFAR-10 test set: one in color ({\sc Color}) and one in grayscale ({\sc Gray}). These two datasets are considered separately, and only the 10-way classification decision boundary is relevant. 

\smallsec{Discussion} We point out upfront that the analogy between color/grayscale and gender domains here wears thin: (1) we consider the two color/grayscale domains as purely binary and disjoint whereas the concept of gender is more fluid; (2) a color/grayscale domain classifier is significantly simpler to construct than a gender recognition model; (3) the transformation between color and grayscale images is linear whereas the manifestation of gender is much more complex. 

Nevertheless, we adopt this simple framework to distill down the core algorithmic exploration before diving into the more complex setups in Sec.~\ref{sec:real-world}. This formulation has several compelling properties: (1) we can control the correlation synthetically by changing images from color to grayscale, maintaining control over the distribution, (2) we can guarantee that color images contain strictly more information than grayscale images, maintaining control over the discriminative cues in the images, and (3) unlike other datasets, there is no fairness/accuracy trade off since both are complementary. Furthermore, despite its simplicity, this setup still allows us to study the behavior of modern CNN architectures.

\smallsec{Key Issue} We ground the discussion by presenting one key result that is counter-intuitive and illustrates why this very simple setting is reflective of a much deeper problem. We train a standard ResNet-18~\cite{he_deep_2015} architecture with a softmax and cross-entropy loss for 10-way object classification. Training on the skewed CIFAR-10S dataset and testing on {\sc Color} images yields $89.0\pm 0.5\%$ accuracy.\footnote{We report the mean across 5 training runs (except for CelebA in Sec.~\ref{sec:real-world-celeba}). Error bars are 2 standard deviations (95\% confidence interval).} This may seem like a reasonable result until we examine that a model trained on an all-grayscale training set (so never having seen a single color image!) yields a significantly higher $93.0\%$ accuracy when tested out-of-domain on {\sc Color} images. 

\begin{figure}
    \centering
    \includegraphics[width=.6\linewidth]{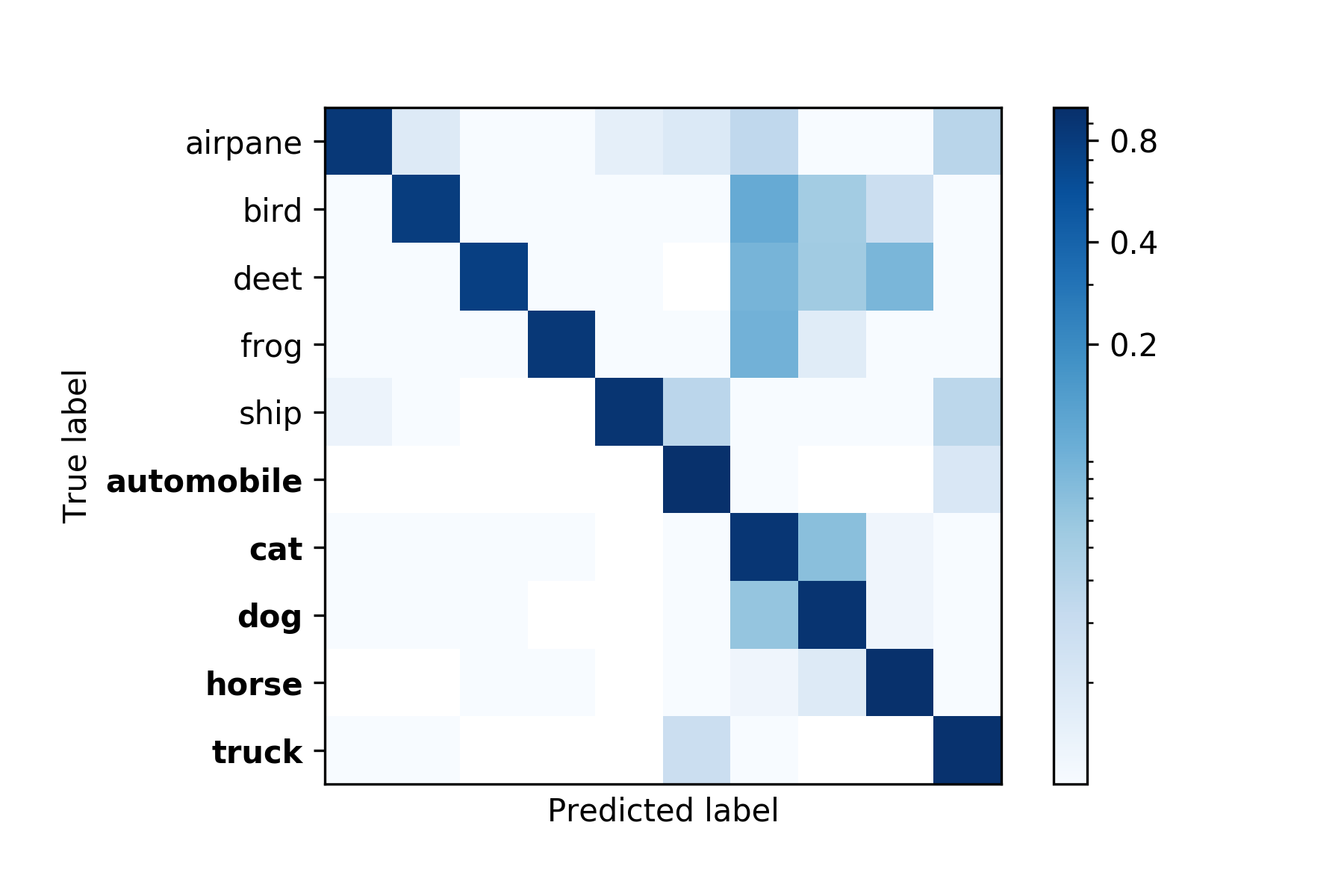}
    %\vspace{-0.1in}
    \caption{Confusion matrix of a ResNet-18~\cite{he_deep_2015} classifier trained on the skewed CIFAR-10S dataset.  The model has learned to correlate the presence of color with the five object classes (in bold) and predominantly predicts those classes on the all-color test set.}
    %\vspace{-0.1in}
    \label{fig:confusion}
\end{figure}

This disparity occurs because the model trained on CIFAR-10S learned to correlate the presence of color and the object classes. When faced with an all-color test set, it infers that it is likely that these images come from one of the five classes that were predominantly colored during training (Fig.~\ref{fig:confusion}). In a real world bias setting where the two domains correspond to gender and the classification targets correspond to activities, this may manifest itself as the model making overly confident predictions of activities traditionally associated with female roles on images of women~\cite{zhao_men_2017}.

\section{Benchmarking Bias Mitigation Methods}
\label{sec:cifar-baselines}

Grounded with the task at hand (training recognition models in the presence of spurious correlations) we perform a thorough benchmark evaluation of bias mitigation methods. Many of these techniques have been proposed in the literature for this task; notable exceptions include  prior shift inference for bias mitigation (Sec.~\ref{sec:cifar-baselines-discriminative}), the distinction between discriminative and conditional training in this context (Sec.~\ref{sec:cifar-baselines-independent}), and the different inference methods for conditional training from biased data (Sec.~\ref{sec:cifar-baselines-independent}). Our findings are summarized in Table~\ref{table:cifar}. In Sec.~\ref{sec:real-world} we demonstrate how our findings on CIFAR10S generalize to real world settings. 

\begin{table*}[t]
\centering
\begin{footnotesize}
\begin{tabular}{l?l?l?c?ccc}
\toprule
&&&& \multicolumn{3}{c}{\sc Accuracy (\%, $\uparrow$)}\\
{\sc Model Name} & {\sc Model} & {\sc Test Inference} & {\sc Bias ($\downarrow$)} & {\sc Color} & {\sc Gray} & {\sc Mean} \\
\midrule
{\sc Baseline} & N-way softmax & $\argmax_y \mathrm{P}(y|x)$ & $0.074$ & $89.0$ & $88.0$ & $88.5 \pm 0.3$\\
{\sc Oversampling} & N-way softmax, resampled & $\argmax_y \mathrm{P}(y|x)$ & $0.066$ & $89.2$ & $89.1$ & $89.1\pm0.4$\\
\midrule
\multirow{2}{*}{{\sc Adversarial}} & w/ uniform confusion~\cite{alvi2018turning,tzeng2015simultaneous}  & $\argmax_y \mathrm{P}(y|x)$ & $0.101$ &
$83.8$ & $83.9$ & $83.8\pm1.1$ \\
& w/ \(\nabla\) reversal, proj. \cite{zhang2018mitigating}  & $\argmax_y \mathrm{P}(y|x)$ & $0.094$ &
$84.6$ & $83.5$ & $84.1\pm1.0$\\
\midrule 
\multirow{4}{*}{{\sc DomainDiscrim}} & \multirow{3}{*}{joint ND-way softmax}  
&  $\argmax_y \sum_d \ptr(y,d|x)$ & $0.844$&$88.3$ & $86.4$ & $87.3\pm0.3$ \\ 
& & $\argmax_{y} \max_d \pte(y,d|x)$ & $0.040$ &$91.3$ & $89.3$ & $90.3\pm0.5$ \\ 
& & $\argmax_y \sum_d \pte(y,d|x)$ & $0.040$&
$91.2$ & $89.4$ & $90.3\pm0.5$ \\
\cmidrule{2-7}
& RBA \cite{zhao_men_2017} & \( y = \Lagr(\sum_d \ptr(y, d | x)) \) & $0.054$ & $89.2$ & $88.0$ & $88.6\pm0.4$ \\
\midrule
\multirow{2}{*}{{\sc DomainIndepend}} & \multirow{2}{*}{N-way classifier per domain} & 
 $\argmax_y \pte(y|\knownd, x)$ & $0.069$ & $89.2$ & $88.7$ & $88.9\pm0.4$ \\
&&$\argmax_y \sum_d s(y,d,x)$ & $\mathbf{0.004}$ &$\mathbf{92.4}$ & $\mathbf{91.7}$ & $\mathbf{92.0\pm0.1}$ \\
\bottomrule
\end{tabular}
\end{footnotesize}
\vspace{0.1in}
\caption{Performance comparison of algorithms on CIFAR-10S. All architectures are based on ResNet-18 \cite{he_deep_2015}. We investigate multiple bias mitigation strategies, and demonstrate that a domain-independent classifier outperforms all baselines on this benchmark.
}
\label{table:cifar}
%\vspace{-0.05in}
\end{table*}

\smallsec{Setup} To perform this analysis, we utilize the CIFAR-10S domain correlation benchmark of Sec.~\ref{sec:cifar}. We assume that at training time the domain labels are available (e.g., we know which images are color and which are grayscale in CIFAR-10S, or which images correspond to pictures of men or women in the real-world setting). All experiments in this section build on the ResNet-18~\cite{he_deep_2015} architecture trained on the CIFAR-10S dataset, with $N=10$ object classes and $D=\{\mathrm{color}, \> \mathrm{grayscale}\}$. The models are trained from scratch on the target data, removing any potential effects from pretraining. Unless otherwise noted the models are tarined for \( 200 \) epochs, with SGD at a learning rate of \( 10^{-1}\) with a factor of \( 10 \) drop-off every \( 50 \) epochs, a weight decay of \( 5 \mathrm{e}{-4} \), and a momentum of 0.9. During training, the image is padded with 4 pixels on each side and then a $32\times32$ crop is randomly sampled from the image or its horizontal flip.

\smallsec{Evaluation} We consider two metrics: mean per-class per-domain accuracy (primary) and bias amplification of~\cite{zhao_men_2017}. The test set is fully balanced across domains, so mean accuracy directly correlates with the model's ability to avoid learning the domain correlation during training. We include the mean bias metric for completeness with the literature, as

\begin{equation}
\label{eq:bias}
\frac{1}{|C|}\sum_{c \in C} \frac{\max(\mathrm{Gr}_c, \mathrm{Col}_c)}{\mathrm{Gr}_c + \mathrm{Col}_c} - 0.5.
\end{equation}

where $\mathrm{Gr}_c$ is the number of grayscale test set examples predicted to be of class $c$, while $\mathrm{Col}_c$ is the same for color.

%%%%%%%%%%%%%%%%%%%%%%%%%%%%%%%%%%%%%%%%%%%%%%%%%%%%%%%%%%
%%%%%%%%%%%%%%%%%%%%%%%%%%%%%%%%%%%%%%%%%%%%%%%%%%%%%%%%%%
%%%%%% OVERSAMPLING
%%%%%%%%%%%%%%%%%%%%%%%%%%%%%%%%%%%%%%%%%%%%%%%%%%%%%%%%%%
%%%%%%%%%%%%%%%%%%%%%%%%%%%%%%%%%%%%%%%%%%%%%%%%%%%%%%%%%%

\subsection{Strategic Sampling}
\label{sec:cifar-baselines-sampling}
The simplest approach is to strategically sample with replacement to make the training data `look' balanced with respect to the class-domain frequencies. That is, we sample rare examples more often during training, or, equivalently, utilize non-uniform misclassification cost~\cite{Elkan01thefoundations,BickelJMLR09}. However, as detailed in~\cite{weiss2007cost}, there are significant drawbacks to oversampling: (1) seeing exact copies of the same example during training makes overfitting likely, (2) oversampling increases the number of training examples without increasing the amount of information, which increases learning time.

\smallsec{Experimental Evaluation} The baseline model first presented in Sec.~\ref{sec:cifar} is a ResNet-18 CNN with a softmax classication layer, which achieves $88.5\pm0.3\%$ accuracy. The same model with oversampling improves to $89.1\pm0.4\%$ accuracy. Both models drive the training loss to zero. Note that data augmentation is critical for this result: without data augmentation the oversampling model achieves only $79.2\pm0.8\%$ accuracy, overfitting to the data. 

%%%%%%%%%%%%%%%%%%%%%%%%%%%%%%%%%%%%%%%%%%%%%%%%%%%%%%%%%%
%%%%%%%%%%%%%%%%%%%%%%%%%%%%%%%%%%%%%%%%%%%%%%%%%%%%%%%%%%
%%%%%% ADVERSARIAL
%%%%%%%%%%%%%%%%%%%%%%%%%%%%%%%%%%%%%%%%%%%%%%%%%%%%%%%%%%
%%%%%%%%%%%%%%%%%%%%%%%%%%%%%%%%%%%%%%%%%%%%%%%%%%%%%%%%%%
\subsection{Adversarial Training}
\label{sec:cifar-baselines-adversarial}

 Another approach to bias mitigation commonly suggested in the literature is \textit{fairness through blindness}. That is, if a model does not look at, or specifically encode, information about a protected variable, then it cannot be biased. To this end, adversarial training is set up through the minimax objective: maximize the classifier's ability to predict the class, while minimizing the adversary's ability to predict the protected variable based on the underlying learned features. 
 
This intuitive approach, however, has a major drawback. Suppose we aim to have equivalent feature representations across domains. Even if a particular protected attribute does not exist in the feature representation of a classifier, combinations of other attributes can be used as a proxy. This phenomenon is termed \emph{redundant encoding} in the literature \cite{hardt2016equality, dwork2012fairness}. For an illustrative example, consider a real-world task of a bank evaluating a loan application, irrespective of the applicant's gender. Suppose that the applicant's employment history lists `nurse'. It can thus, by proxy, be inferred with high probability that the applicant is also a woman. However, employment history is crucial to the evaluation of a loan application, and thus the removal of this redundant encoding will degrade its ability to perform the evaluation. 

\begin{figure}
	\centering
	\begin{tabular}{c@{}c@{}|@{}c@{}c}
		\multicolumn{2}{c}{\small w/o adversary} & \multicolumn{2}{c}{\small w/ adversary} \\
		{\small domains} & {\small classes} & {\small domains} & {\small classes} \\
		\includegraphics[width=0.24\linewidth]{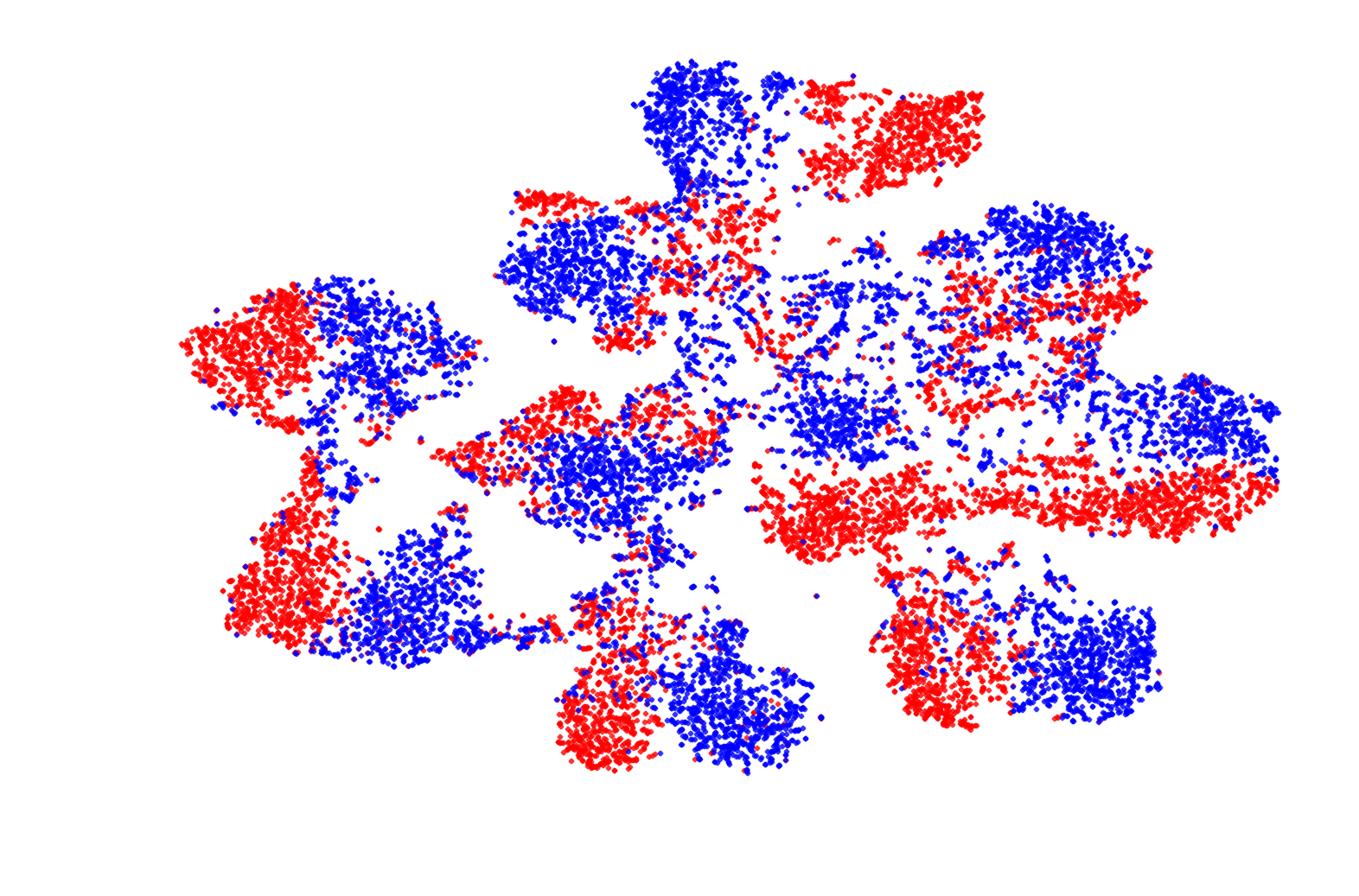} & 
		\includegraphics[width=0.24\linewidth]{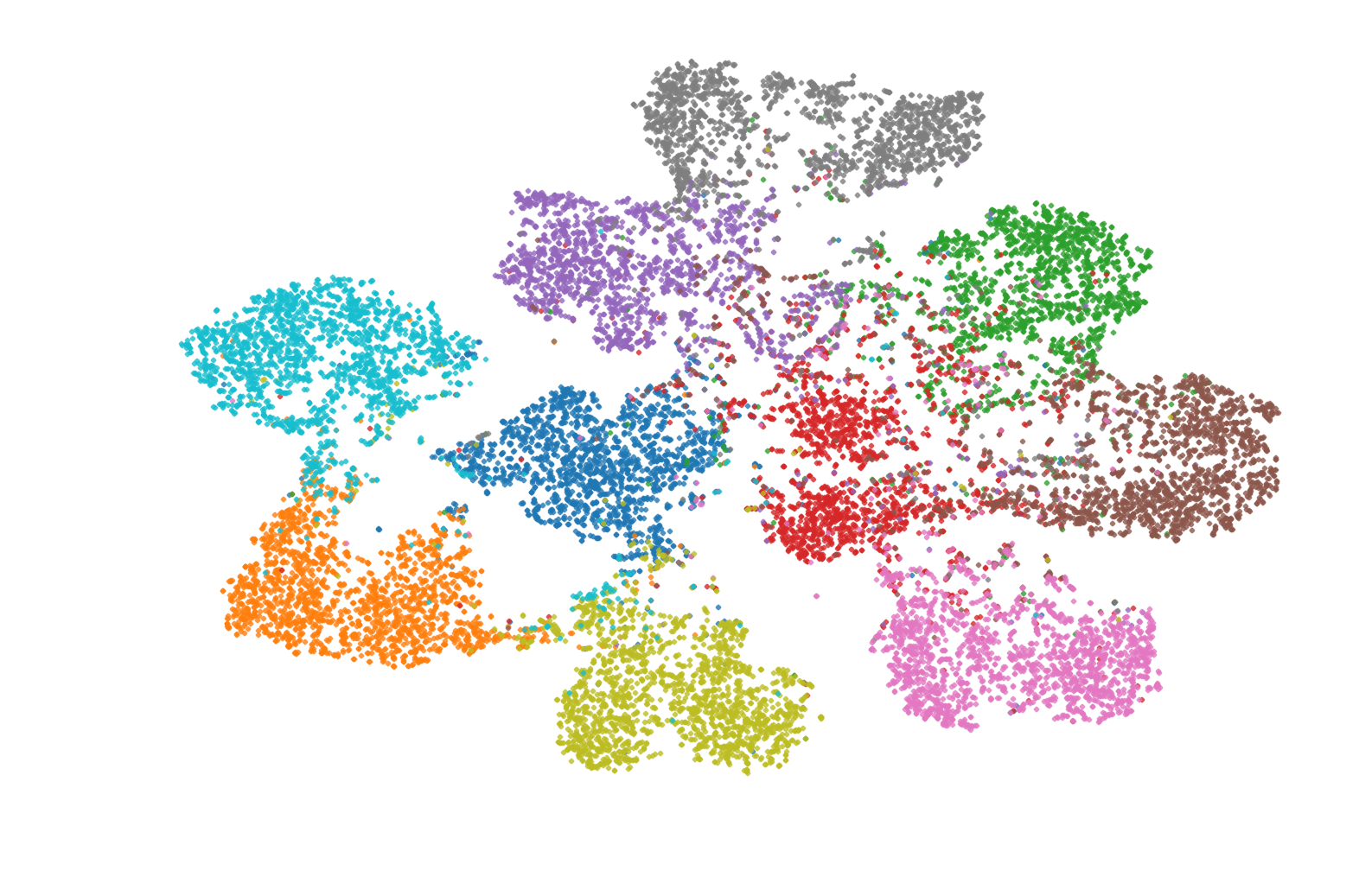} & 
		\includegraphics[width=0.24\linewidth]{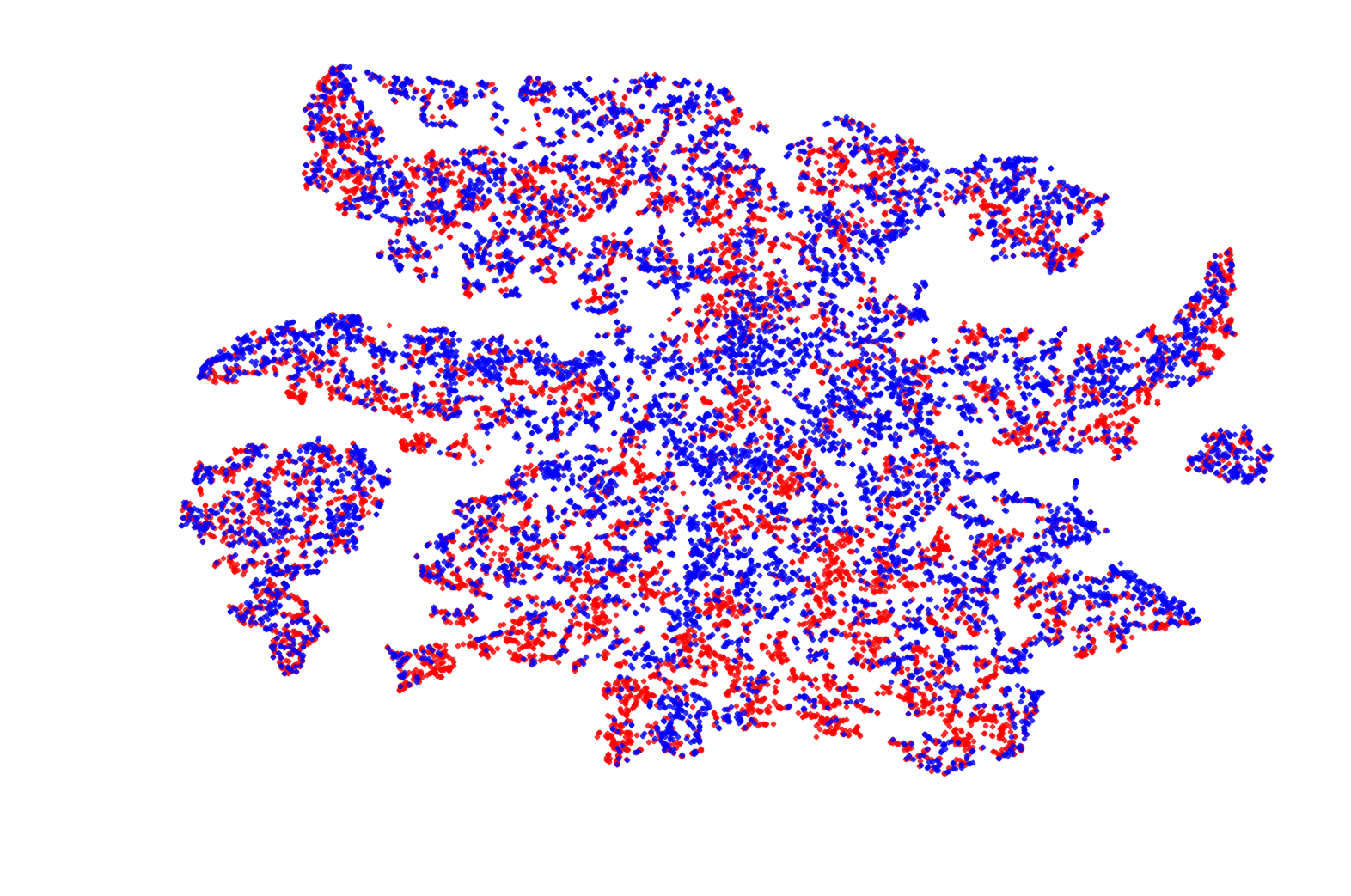} &
		\includegraphics[width=0.24\linewidth]{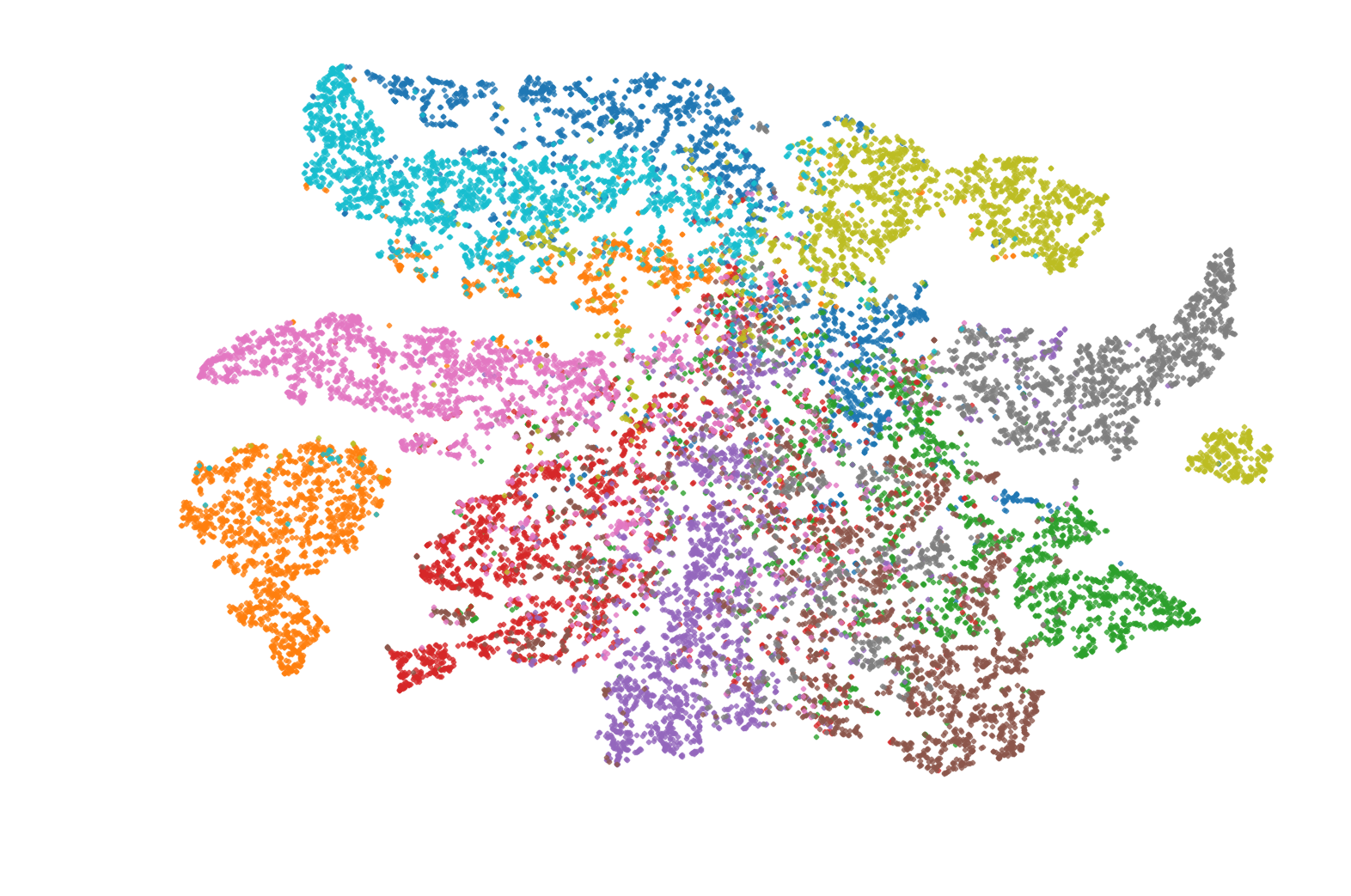} \\
	\end{tabular}
	
	\caption{Adversarial training \cite{zhang2018mitigating} enforces domain confusion but also introduces unwanted class boundary confusion (t-SNE plots).} 
	\vspace{-0.1in}
	\label{fig:adversarial}
\end{figure}

\smallsec{Experimental Evaluation} We apply adversarial learning to de-bias the object classifier. We consider both the uniform confusion loss \(-(1/|D|) \sum_{d} \log q_d \) of~\cite{alvi2018turning} (inspired by~\cite{tzeng2015simultaneous}), and the loss reversal \( \sum_{d} \mathds{1}[\widehat{d} = d] \log q_d \) with gradient projection of~\cite{zhang2018mitigating}.\footnote{We apply the adversarial classifiers on the penultimate layer for~\cite{alvi2018turning,tzeng2015simultaneous} model, and on the final classification layer for~\cite{zhang2018mitigating} as recommended by the authors. We experimented with other combinations of layers and losses, including applying the projection method of~\cite{zhang2018mitigating} onto the confusion loss of~\cite{alvi2018turning,tzeng2015simultaneous}, and achieved similar results. 
The models are trained for \( 500 \) epochs using Adam with learning rates 3e-4 and weight decay 1e-4. We hold out 10,000 images to tune the hyperparameters before retraining the network on the entire training set. To verify training efficacy, we train SVM domain classifiers on the learned features:  the accuracy is \( 99.0\% \) before and \( 78.2\% \) after adversarial training, verifying training effectiveness.} These methods achieve only $83.4\%$ and $84.1\%$ accuracy, respectively. As Fig.~\ref{fig:adversarial} visually demonstrates, although the adversarial classifier enforces domain confusion it additionally creates undesirable class confusion. 

We run one additional experiment to validate the findings. We test whether models encode the domain (color/grayscale) information even when \emph{not} exposed to a biased training distribution; if so, this would help explain why minimizing this adversarial objective would lead to a worse underlying feature representation and thus reduced classification accuracy. We take the feature representation of a 10-way classifier trained on \emph{all color} images (so not exposed to color/grayscale skew) and train a linear SVM adversary on this feature representation to predict the color/grayscale domain of a new image. This yields an impressive 82\% accuracy; since the ability to discriminate between the two domains emerges naturally even without biased training, it would make sense that requiring that the model not be able to distinguish between the two domains would harm its overall classification ability.

%%%%%%%%%%%%%%%%%%%%%%%%%%%%%%%%%%%%%%%%%%%%%%%%%%%%%%%%%%
%%%%%%%%%%%%%%%%%%%%%%%%%%%%%%%%%%%%%%%%%%%%%%%%%%%%%%%%%%
%%%%%% DISCRIMINATIVE
%%%%%%%%%%%%%%%%%%%%%%%%%%%%%%%%%%%%%%%%%%%%%%%%%%%%%%%%%%
%%%%%%%%%%%%%%%%%%%%%%%%%%%%%%%%%%%%%%%%%%%%%%%%%%%%%%%%%%

\subsection{Domain Discriminative Training}

\label{sec:cifar-baselines-discriminative}
The alternative to fairness through blindness is \emph{fairness through awareness}~\cite{dwork2012fairness} where the domain information is first explicitly encoded and then explicitly mitigated. The simplest approach is training a $ND$-way discriminative classifier where $N$ is the number of target classes and $D$ is the number of domains. The correlation between domains and classes can then be removed during inference in one of several ways.

\subsubsection{Prior Shift Inference}

If the outputs of the $ND$-way classifier can be interpreted as probabilities, a test-time domain solution to removing class-domain correlation was introduced in~\cite{saerens2002adjusting} and applied in~\cite{royer2015classifier} to visual recognition.  Let the classifier output a joint probability $\pr(y,d|x)$ for target class $y$, domain $d$ and image $x$. We can assume that $\ptr(x|y,d)=\pte(x|y,d)$, i.e., the distribution of image appearance within a particular class and domain is the same between training and test time. However, $\ptr(d,y)\neq\pte(d,y)$, i.e., the correlation between target classes and domains may have changed. This suggests that the test-time probability $\pte(y,d|x)$ should be computed as:

\begin{align}
    \pte(y,d | x) &\propto \pte(x | y,d)\pte(y,d) \label{eq:ps:1}\\
    &=\ptr(x|y,d)\pte(y,d) \label{eq:ps:2}\\
    &\propto \ptr(y,d|x)\frac{\pte(y,d)}{\ptr(y,d)} \label{eq:ps:3} 
\end{align}

In theory, this requires access to the test label distribution $\pte(y,d)$; however, assuming uncorrelated $d$ and $y$ at test time (unbiased $\pte(d|y)$) and mean per-class accuracy evaluation (uniform $\pte(y)$), $\pte(y,d)=\pte(d|y)\pte(y)\propto1$.

Eqn.~\ref{eq:ps:3} then simplifies to $\ptr(y,d|x)/\ptr(y,d)$, removing the test distribution requirement. With this assumption, the target class predictions can be computed directly as

\begin{equation}
    \hat{y} = \argmax_y \max_d \pte(y,d|x)
    \label{eq:discrmax}
\end{equation}

or, using the Law of Total Probability, 
\begin{equation}
\label{eq:discrsum}
    \hat{y} = \argmax_y \pte(y|x)=\argmax_y \sum_d\pte(y,d|x).
\end{equation}

\smallsec{Experimental Evaluation} We train a $ND$-way classifier (20-way softmax in our setting) to discriminate between (class, domain) pairs. This discriminative model with inference prior shift towards a uniform test distribution (Eqn.~\ref{eq:ps:3}) followed by sum of outputs (Eqn.~\ref{eq:discrsum}) achieves $90.3\%$ accuracy,  significantly outperforming the $88.5\pm0.3\%$ accuracy of the $N$-way softmax baseline. To quantify the effects of the two steps of inference: taking the highest output predictor rather than summing across domains (Eqn.~\ref{eq:discrmax}) has no effect on accuracy because the two domains are easily distinguishable in this case; however, summing the outputs without first applying prior shift drops accuracy from $90.3\%$ to $87.3\%$.

Finally, we verify that the increase in accuracy is not just the result of the increased number of parameters in the classifier layer. We train an ensemble of baseline models, averaging their softmax predictions: one baseline achieves $88.5\%$ accuracy, two models achieve $89.6\%$, and only an ensemble of \emph{five} baseline models (with 55.9M trainable parameters) achieve $90.0\%$ accuracy on par with $90.3\%$ accuracy of the discriminative model (with 11.2M parameters).

%%%%%%%%%%%%%%%%%%%%%%%%%%%%%%%%%%%%%%%%%%%%%%%%%%%%%%%%%%
%%%%%%%%%%%%%%%%%%%%%%%%%%%%%%%%%%%%%%%%%%%%%%%%%%%%%%%%%%
%%%%%% LAGRANGIAN
%%%%%%%%%%%%%%%%%%%%%%%%%%%%%%%%%%%%%%%%%%%%%%%%%%%%%%%%%%
%%%%%%%%%%%%%%%%%%%%%%%%%%%%%%%%%%%%%%%%%%%%%%%%%%%%%%%%%%
\vspace{-0.1in}
\subsubsection{Reducing Bias Amplification}

 An alternative inference approach is Reducing Bias Amplification (``RBA'') of Zhao et al.~\cite{zhao_men_2017}. RBA uses corpus-level constraints to ensure inference predictions follow a particular distribution. They propose a Lagrangian relaxation iterative solver since the combinatorial optimization problem is challenging to solve exactly at large scale. This method effectively matches the desired inference distribution and reduces bias; however, the expensive optimization must be run on all test samples before a single inference is possible. 

\smallsec{Experimental Evaluation} 
In the original setting of \cite{zhao_men_2017}, training and test time biases are equal. However, RBA is flexible enough to optimize for any target distribution. On CIFAR-10S, we thus set the optimization target bias to 0 and the constraint epsilon to \(5\%\). To make the optimization as effective as possible, we substitute in the known test-time domain (because it can be perfectly predicted) so that the optimization only updates the class predictions.

Applying RBA on the $\sum_d \ptr(y, d |x )$ scores results in $88.6\%$ accuracy, a $1.3\%$ improvement over the simpler $\argmax_y \sum_d \ptr(y, d |x )$ inference but an insignificant improvement over $88.5\%$ of the {\sc Baseline} model. Interestingly, we also observe that the benefits of RBA optimization are significantly lessened when prior shift is applied beforehand. For example, when using the $\sum_d \pte(y, d |x )$ post-prior shift scores, accuracy only improves negligibly from $90.3\%$ using $\argmax$ inference to $90.4\%$ using RBA. Therefore, we conclude that applying RBA after prior shift is extraneous. However, the converse is not true as the best accuracy achieved by RBA without prior shift is significantly lower than the accuracy achieved with prior shift inference. 

\subsection{Domain Independent Training}
\label{sec:cifar-baselines-independent}

One concern with the discriminative model is that it learns to distinguish between the $ND$ class-domain case; in particular, it explicitly learns the boundary between the same class across different domains (e.g., cat in grayscale versus cat in color, or a woman programming versus a man programming). This may be wasteful, as the $N$-way class decision boundaries may in fact be similar across domains and the additional distinction between the same class in different domains may not be necessary. Furthermore, the model is necessarily penalized in cases where the domain prediction is challenging but the target class prediction is unambiguous. 

This suggests training separate classifiers per domain. Doing this naively, however, as an ensemble, will yield poor performance as each model will only see a fraction of the data. We thus consider  a shared feature representation with an ensemble of classifiers. This alleviates the data reduction problem for the representation though not for the classifiers.

Given the predictions $\pr(y|d,x)$, multiple inference methods are possible. If the domain $\knownd$ is known at test time, $\hat{y} = \argmax_y \pr(y|\knownd,x)$ is reasonable yet entirely ignores the learned class boundaries in the other domains $d \neq \knownd$, and may suffer if some classes $y$ were poorly represented within $\knownd$ during training. If a probabilistic interpretation is possible, then two inference methods are reasonable:
\begin{align}
    \hat{y} &= \argmax_y \max_d \pr(y|d,x), \text{ or } \label{eq:condinf:1}\\
    \hat{y} &= \argmax_y \sum_d \pr(y|d,x)\pr(d|x) \label{eq:condinf:2}
\end{align}

However, Eqn.~\ref{eq:condinf:1} again ignores the learned class boundaries across domains, and Eqn.~\ref{eq:condinf:2} requires inferring $\pr(d|x)$ (which may either be trivial, as in CIFAR-10S, reducing to a single-domain model, or complicated to learn and implicitly encoding the correlations between $y$ and $d$ that we are trying to avoid). Further, in practice, while the probabilistic interpretation of a single model may be a reasonable approximation, the probabilistic outputs of the multiple independent models are frequently miscalibrated with respect to each other. 

A natural option is to instead reason directly on class  boundaries of the $D$ domains, and perform inference as\footnote{Interestingly, under a softmax probabilistic model this inference corresponds to the geometric mean between $\{\pr(y|d,x)\}_d$, which is a stable method for combining independent models with different output ranges. }
\begin{equation}
\label{eq:condinf:3}
    \hat{y} = \argmax_y \sum_d \score(y,d,x),
\end{equation}
where \( s(y, d, x) \) are the network activations at the classifier layer. For linear classifiers with a shared feature representation this corresponds to averaging the class decision boundaries. We demonstrate that this technique works  well in practice across both single and multi-label target classification tasks at removing class-domain correlations. 

\smallsec{Experimental Evaluation} We train a model for performing object classification on the two domains independently. This is implemented as two 10-way independent softmax classifiers sharing the same underlying network. At training time we use knowledge of the image domain to only update one of the classifiers. At test time we apply prior shift to adjust the output probabilities of both classifiers towards a uniform distribution, and consider two inference methods. First, we use only the classifier corresponding to the test domain, yielding a low $88.9\%$ accuracy as expected because it is not able to integrate information across the two domains (despite requiring specialized knowledge of the image domain). Instead, we combine the decision boundaries following Eqn.~\ref{eq:condinf:3} and achieve $92.0\%$ accuracy, significantly outperforming the baseline of $88.5 \pm 0.3\%$. 

\vspace{-0.01in}  
\subsection{Summary of Findings}
\vspace{-0.01in}
So far we illustrated that the CIFAR-10S setup is an effective benchmark for studying  bias mitigation, and provided a thorough evaluation of multiple techniques. We demonstrated the shortcomings of strategic resampling and of adversarial approaches for bias mitigation. We showed that the prior shift inference adjustment of output probabilities is a simpler, more efficient, and more effective alternative to the RBA technique~\cite{zhao_men_2017}.   Finally, we concluded that the domain-conditional model with explicit combination of per-domain class predictions significantly outperforms all other techniques.  Table~\ref{table:cifar} lays out the findings.

Recall our original goal of Sec.~\ref{sec:cifar} to train a model that mitigates the domain correlation bias in CIFAR-10S enough to classify color images of objects as well as a model trained on only grayscale images would. We have partially achieved that goal. The {\sc DomainIndependent} model trained on CIFAR-10S achieves $92.4\%$ accuracy on color images, significantly better than $89.0\pm0.5\%$ of {\sc Baseline} and approaching $93.0\pm0.2\%$ of the model trained entirely on grayscale images. However, much still remains to be done. We would expect that a model trained on CIFAR-10S would take advantage of the available color cues and perform even better than $93.0\%$, ideally approaching $95.1\%$ accuracy of a model trained on all color images. The correlation bias is a much deeper problem for visual classifiers and much more difficult to mitigate than it appears at first glance.

\vspace{-0.01in}
\section{Real World Experiments}
\vspace{-0.01in}
\label{sec:real-world}

While CIFAR-10S proves to be a useful landscape for bias isolation studies, there remains the implicit assumption throughout that such findings will generalize to other settings. Indeed, it is possible that they may not due to the synthetic nature of the proposed bias generation. We thus investigate our findings in three alternative scenarios. First, in Sec.~\ref{sec:real-world-cifar} we consider two modifications to CIFAR-10S: varying the level of skew beyond the 95\%-5\% studied in Sec.~\ref{sec:cifar-baselines}, and replacing the color/grayscale domains with more realistic non-linear transformations. After verifying all our findings still hold, in Sec.~\ref{sec:real-world-celeba} we consider face attribute recognition on the CelebA dataset~\cite{liu2015faceattributes} where the presence of attributes, e.g., ``smiling'' is correlated with gender.

\subsection{CIFAR Extensions}
\label{sec:real-world-cifar}

There are two key distinctions between the CIFAR-10S dataset studied in Sec.~\ref{sec:cifar-baselines} and the real world scenarios where gender or race are correlated with the target outputs.

\begin{figure}[t]
    \centering
    \begin{tabularx}{\linewidth}{m{0.6\linewidth}@{\hskip 0.05in}m{0.3\linewidth}}
       \includegraphics[height=1.2in]{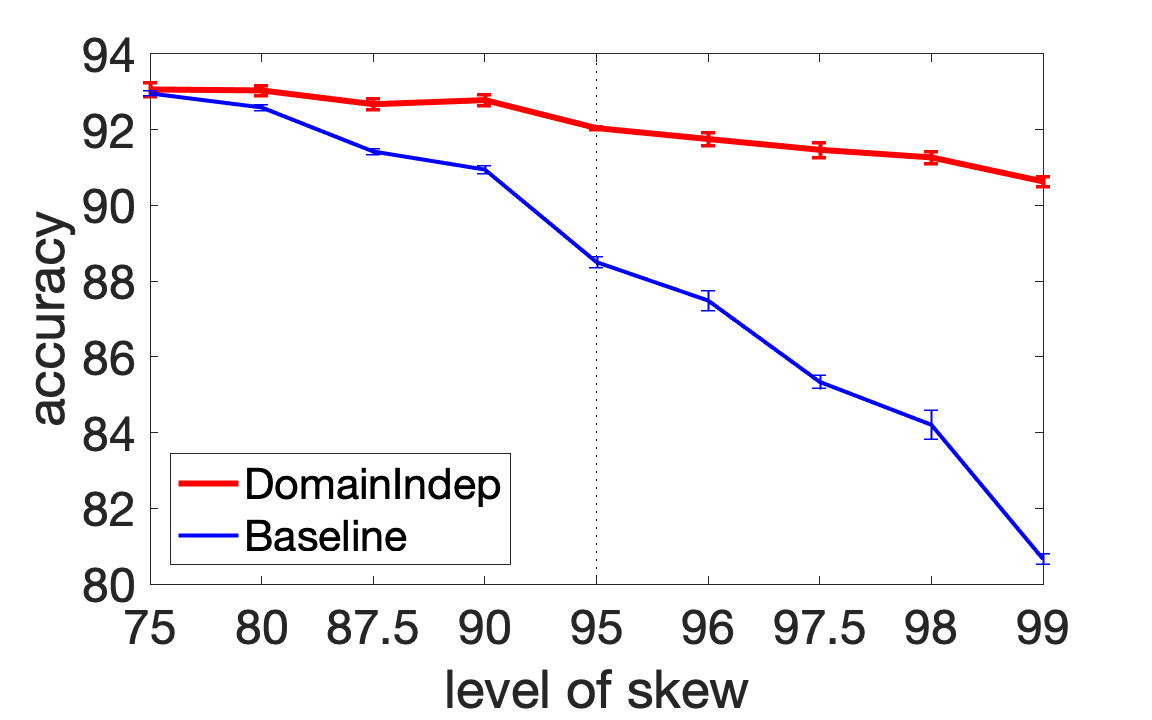}&
       \includegraphics[height=1.2in]{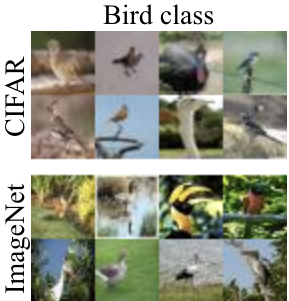} 
    \end{tabularx}
    \vspace{-0.07in}
    \caption{\emph{(Left)} The {\sc DomainIndep} model outperforms the {\sc Baseline} on CIFAR-10S for varying levels of skew. \emph{(Right)} To investigate more real-world domains instead of color-grayscale, we consider the subtle  shift between CIFAR and 32x32 ImageNet~\cite{CINIC,ILSVRC15}. }
    \label{fig:real-world-cifar}
\end{figure}

\smallsec{Varying Degrees of Domain Distribution} The first distinction is in the \emph{level} of skew, where domain balance may be more subtle than the 95\%-5\% breakdown studied above. To simulate this setting, we validated on CIFAR with different levels of color/grayscale skew, using the setup of Sec.~\ref{sec:cifar-baselines} in Fig.~\ref{fig:real-world-cifar} (\emph{Left}). The {\sc DomainIndep} model consistently outperforms the {\sc Baseline}, although the effect is significantly more pronounced at higher skew levels. For reference, the average gender skew on the CelebA dataset~\cite{liu2015faceattributes} for face attribute recognition described in Sec.~\ref{sec:real-world-celeba} is $80.0\%$\footnote{In this multi-label setting the gender skew is computed on the dev set as the mean across 39 attributes of $\frac{\min(|attr=1,woman|,|attr=1,man|)}{|attr=1|}$.}.

\smallsec{Other Non-Linear Transformations} The second distinction is that real-world protected attributes differ from each other in more than just a linear color-grayscale transformation (e.g., men and women performing the same task look significantly more different than the same image in color or grayscale). To approximate this in a simple setting, we followed the CIFAR protocol of Sec.~\ref{sec:cifar-baselines}, but instead of converting images to grayscale, we consider alternative domain options in Table~\ref{table:nonlinear}. Arguably the most interesting shift corresponds to taking images of similar classes from ImageNet~\cite{ILSVRC15,CINIC}, and we focus our discussion on that one. 

The domain shift here is subtle (shown in Fig.~\ref{fig:real-world-cifar} \emph{Right}) but the conclusions hold: mean per-class per-domain accuracy is {\sc Baseline} $79.4\pm0.4\%$, {\sc Adversarial} $74.1\pm0.6$\%~\cite{alvi2018turning,tzeng2015simultaneous} and $73.1\pm3.0$\%~\cite{zhang2018mitigating} (not shown in Table~\ref{table:nonlinear}), {\sc DomainDiscriminative} $81.5\pm 0.7\%$, and our {\sc DomainIndependent} model $83.5\pm0.3\%$.
One interesting change is that {\sc Oversampling}  yields $78.6\pm0.4\%$, significantly lower than the baseline of $79.4\%$, so we investigate further. The drop can be explained by the five classes which were heavily skewed towards CIFAR images at training time: the model overfit to the small handful of ImageNet images which got oversampled, highlighting the concerns with oversampling particularly in situations where the two domains are different from each other and the level of imbalance is high. We observe similar results in the high-to-low-resolution domain shift (third and fourth columns of Table~\ref{table:nonlinear}), where the two domains are again very different from each other. To counteract this effect we instead applied the class-balanced loss method Cui et al.~\cite{Cui_2019_CVPR}, cross-validating the hyperparameter on a validation set to $\beta = 0.9$, and achieved a more reasonable result of $79.2\%$, on par with $79.4\pm0.4\%$ of {\sc Baseline} but still behind $83.5\pm0.3\%$ of {\sc DomainIndependent}.

\begin{table}[t]
\centering
\begin{footnotesize}
\begin{tabular}{l?cccc}
\toprule
{\sc Model} & 28x28crop & 1/2 res. & 1/4 res. &  ImageNet\\
\midrule
{\sc Baseline} & 89.2 & 85.6 & 73.7  & 79.4 \\
{\sc Oversamp} & 90.1 & 85.4 & 72.7  & 78.6 \\
{\sc DomDiscr} & 91.6 & 88.5 & 77.3  & 81.5 \\
{\sc DomIndep} & {\bf 93.0} & {\bf 90.2} & {\bf 79.9}  & {\bf 83.5} \\
\bottomrule
\end{tabular}
\end{footnotesize}
\vspace{0.1in}
\caption{On CIFAR-10S, we consider other transformations instead of the grayscale domain: (1) cropping the center of the image, (2,3) reducing the image resolution~\cite{su2017adapting}, followed by upsampling or (4) replacing with 32x32 ImageNet images of the same class~\cite{CINIC}. We use the inference of Eqn.~\ref{eq:discrsum} for {\sc DomDiscr} and Eqn.~\ref{eq:condinf:3} for {\sc DomIndep}, and report mean per-class per-domain accuracy (in \%). Our conclusions from Sec.~\ref{sec:cifar-baselines} hold across all domain shifts. }
\label{table:nonlinear}
\end{table}

\vspace{-0.02in}
\subsection{CelebA Attribute Recognition}
\vspace{-0.01in}
\label{sec:real-world-celeba}

Finally, we verified our findings on the real-world CelebA dataset~\cite{liu2015faceattributes}, used in~\cite{ryu2017improving} to study face attribute recognition when the presence of attributes, e.g., ``smiling,'' is correlated with gender. We trained models to recognize the 39 attributes (all except the ``Male'' attribute). Out of the 39 attributes, 21 occur more frequently with women and 18 with men, with an average gender skew of 80.0\% when an attribute is present.  During evaluation we consider the 34 attributes that have sufficient validation and test images.\footnote{The removed attributes did not contain at least 1 positive male, positive female, negative male, and negative female image. They are: 5 o'clock shadow, goatee, mustache, sideburns and wearing necktie. } 

\smallsec{Task and Metric} The target task is multi-label classification, evaluated using mean average precision (mAP) across attributes. We remove the gender bias in the test set by using a weighted mAP metric: for an attribute that appears with $N_m$ men and $N_w$ women images, we weight every positive man image by $(N_m+N_w)/(2N_m)$ and every positive woman image by $(N_m+N_w)/(2N_w)$ when computing the true positive predictions. This simulates the setting where the total weight of positive examples within the class remains constant but is now equally distributed between the genders.

We also evaluate the bias amplification (BA) of each attribute~\cite{zhao_men_2017}. For an attribute that appears more frequently with women, this is $P_w/(P_m+P_w)-N_w/(N_m+N_w)$ where $P_w,P_m$ are the number of women and men images respectively classified as positive for this attribute. For attributes that appear more frequently with men, the numerators are $P_m$ and $N_m$. To determine the binary classifier decision we compute a score threshold for each attribute which maximizes the classifier's F-score on the validation set. Since our methods aim to de-correlate gender with the attribute we expect that bias amplification will be \emph{negative} as the predictions approach a uniform distribution across genders. 

\smallsec{Training Setup} The images are the Aligned\&Cropped subset of CelebA~\cite{liu2015faceattributes}. We use a ResNet-50~\cite{he_deep_2015} base architecture pre-trained on ImageNet~\cite{ILSVRC15}. The FC layer of the ResNet model is replaced with two consecutive fully connected layers. Dropout and relu is applied to the output between the two fully connected layers, which has size 2048.  It is trained with a binary cross entropy loss with logits using a batch size of 32, for 50 epochs with the Adam optimizer~\cite{adam} (learning rate 1e-4). The best model over all epochs is selected per inference method on the validation set. For adversarial training, we run an extensive hyperparameter search over the relative weights of the losses and the number of epochs of the adversary. We select the model with the highest weighted mAP on the validation set among all models that successfully train a de-biased representation (accuracy of the gender classifier drops by at least 1\%; otherwise it's essentially the {\sc Baseline} model with the same mAP). The models are evaluated on the test set. 

\begin{table}[t]
    \centering
    \footnotesize
    \begin{tabular}{llcc}
\toprule
{\sc Model} & {\sc Model} & {\sc mAP} & {\sc BA} \\
\hline
{\sc Base}  &  N sigmoids & 74.7 & 0.010\\
{\sc Adver} %& uniform confusion [1, 38] & 40.6 \\
&  w/uniform conf.~\cite{alvi2018turning,tzeng2015simultaneous}

& 71.9 & 0.019\\

\multirow{1}{*}{{\sc DomDis}} & 2N sigm, $\sum_d \mathrm{P_{tr}}(y,d|x)$ & 73.8 & 0.007\\ 
\midrule
\multirow{4}{*}{{\sc DomInd}} & 2N sigmoids, $\mathrm{P_{tr}}(y|d^*,x)$ & 73.8 & 0.009\\
& 2N sigm, $\max_d \mathrm{P_{tr}}(y|d,x)$ & 75.4 & {\bf -0.039}\\
& 2N sigm, $\sum_d \mathrm{P_{tr}}(y|d,x)$ & 76.0 & -0.037\\
& 2N sigmoids, $\sum_d s(y,d,x)$ & {\bf 76.3} & -0.035\\
\bottomrule
    \end{tabular}
    \vspace{0.1in}
    \caption{Attribute classification accuracy evaluated using mAP (in \%, $\uparrow$) weighted to ensure an equal distribution of men and women appearing with each attribute, and Bias Amplification ($\downarrow$). Evaluation is on the CelebA test set, across 34 attributes that have sufficient validation data; details in Sec.~\ref{sec:real-world-celeba}. }
    \vspace{-0.1in}
    \label{table:celeba}
\end{table}

\smallsec{Results} Table~\ref{table:celeba} summarizes the results. The overall conclusions from Sec.~\ref{sec:cifar-baselines} hold despite the transition to the multi-label setting and to real-world gender bias. {\sc Adversarial} training as before de-biases the representation but also harms the mAP (71.9\% compared to 74.7\% for {\sc Baseline}). In this multi-label setting we do not consider a probabilistic interpretation of the output as the classifier models are trained independently instead of jointly in a softmax. Without this interpretation and prior shift  the {\sc DomainDiscrminative} model achieves less competitive results than the baseline at $73.8\%$. RBA inference of~\cite{zhao_men_2017} towards a uniform distribution performs similarly at $73.6\%$. The {\sc DomainIndependent} model successfully mitigates gender bias and outperforms the domain-unaware {\sc Baseline} on this task, increasing the weighted mAP from $74.7\%$ to $76.3\%$. Alternative inference methods, such as selecting the known domain, computing the max output over the domains, or summing the outputs of the probabilities directly achieve similar bias amplification results but perform between $0.3-2.5\%$ mAP worse. 

\begin{figure}
    \centering
    \includegraphics[width=0.8\linewidth]{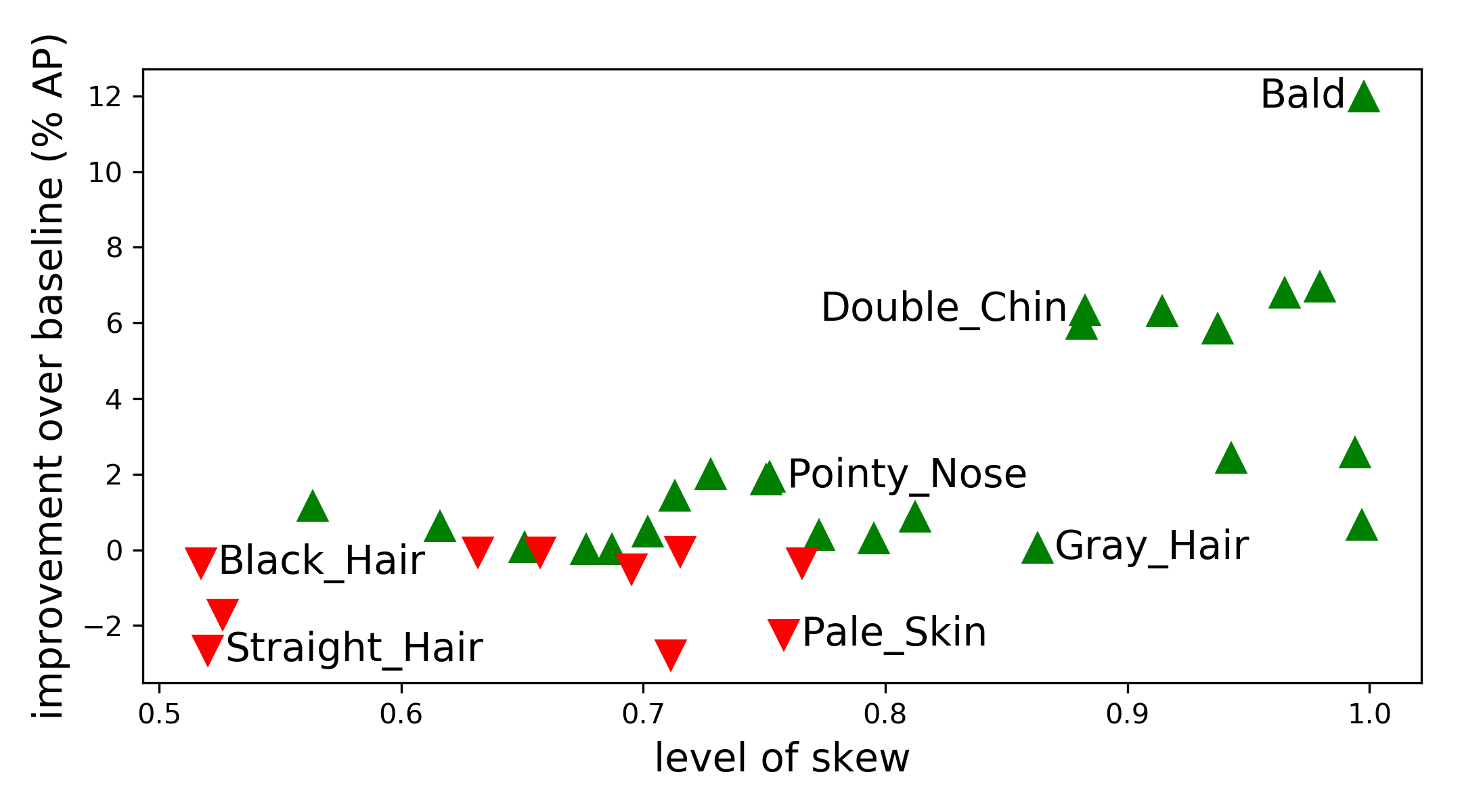}
    \vspace*{-0.05in}
    \caption{Per-attribute improvement of the {\sc DomainIndependent} model over the {\sc Baseline} model on the CelebA validation set, as a function of the level of gender imbalance in the attribute. Attributes with high skew (such as ``bald'') benefit most significantly. }
    \vspace*{-0.1in}
    \label{fig:celeba_imbalance}
\end{figure}

\smallsec{Analysis} We take a deeper look at the per-class results on the validation set to understand the factors that contribute to the improvement. Overall the {\sc DomainIndependent} model improves over {\sc Baseline} on 24 of the 34 attributes. Fig.~\ref{fig:celeba_imbalance} demonstrates that the level of gender skew in the attribute is highly correlated with the amount of improvement ($\rho=0.709$). Attributes that have skew greater than $80\%$ (out of the positive training images for this attribute at least $80\%$ belong to one of the genders) always benefit from the {\sc DomainIndependent} model. This is consistent with the findings from CIFAR-10S in Fig.~\ref{fig:real-world-cifar}\emph{(Left)}. When the level of skew is insufficiently high the harm from using fewer examples when training the {\sc DomainIndependent} model outweighs the benefit of decomposing the representation. 

\smallsec{Oversampling} Finally, we note that the {\sc Oversampling} model in this case achieves high mAP of 77.6\% and bias amplification of -0.061, outperforming the other techniques. This is expected as we know from prior experiments in Sec.~\ref{sec:cifar-baselines} and \ref{sec:real-world-cifar} that oversampling performs better in settings where the two domains are more similar (color/grayscale, 28x28 vs 32x32 crop) and where the skew is low while the dataset size is large so it wouldn't suffer from overfitting.

\vspace{-0.05in}
\section{Conclusions}
\vspace{-0.02in}
We provide a benchmark and a thorough analysis of bias mitigation techniques in visual recognition models. We draw several important algorithmic conclusions, while also acknowledging that this work does not attempt to tackle many of the underlying ethical fairness questions. What happens if the domain (gender in this case) is non-discrete? What happens if the imbalanced domain distribution is not known at training time -- for example, if the researchers failed to identify the undesired correlation with gender? What happens in downstream tasks where these models may be used to make prediction decisions? We leave these and many other questions to future work. 

\smallsec{Acknowledgements} This work is partially supported by the National Science Foundation under Grant No. 1763642, by Google Cloud, and by the Princeton SEAS Yang Family Innovation award. Thank you to Arvind Narayanan and to members of Princeton's Fairness in AI reading group for great discussions.

{\small
\bibliographystyle{ieee_fullname}
\bibliography{egbib}
}

\end{document}